\documentclass[runningheads]{llncs}
\usepackage{graphicx}

\usepackage{tikz}
\usepackage{comment}
\usepackage{amsmath,amssymb} 
\usepackage{color}
\usepackage{multirow}
\usepackage{MnSymbol}
\usepackage[pagebackref=true,breaklinks=true,letterpaper=true,colorlinks,bookmarks=false]{hyperref}
\makeatletter
\newcommand\figcaption{\def\@captype{figure}\caption}
\newcommand\tabcaption{\def\@captype{table}\caption}
\makeatother

\usepackage[accsupp]{axessibility}  

\usepackage[width=122mm,left=12mm,paperwidth=146mm,height=193mm,top=12mm,paperheight=217mm]{geometry}

\begin{document}
\pagestyle{headings}
\mainmatter
\def\ECCVSubNumber{4754}  

\title{Video Interpolation by Event-driven\\ Anisotropic Adjustment of Optical Flow} 


\titlerunning{Video Interpolation by Event-driven Anisotropic Adjustment of Optical Flow}
%
\author{Song Wu\inst{\star 1}               \and
Kaichao You\thanks{Song Wu and Kaichao You contribute equally to this paper. Work done while Song Wu, Kaichao You, Yang Tian are interns at Huawei.}\inst{2}                 \and
Weihua He\thanks{Weihua He and Ziyang Zhang are corresponding authors.}\inst{2} \and 
Chen Yang\inst{1} \and Yang Tian\inst{2}    \and 
\\Yaoyuan Wang\inst{1}                      \and 
Ziyang Zhang\inst{\star \star 1}            \and 
Jianxing Liao\inst{1}}
  
\authorrunning{Wu et al.}
\institute{
Advanced Computing and Storage Lab, Huawei Technologies Co. Ltd 
\\ \email{\{wusong5533, yangchen2017.pku\}@gmail.com}
\\
\email{\{wangyaoyuan1,zhangziyang11,liaojianxing\}@huawei.com} \and 
Tsinghua University
\\ \email{\{ykc20,hwh20,tiany20\}@mails.tsinghua.edu.cn} 
}
\maketitle

\begin{abstract}
Video frame interpolation is a challenging task due to the ever-changing real-world scene. Previous methods often calculate the bi-directional optical flows and then predict the intermediate optical flows under the linear motion assumptions, leading to isotropic intermediate flow generation. Follow-up research obtained anisotropic adjustment through estimated higher-order motion information with extra frames. Based on the motion assumptions, their methods are hard to model the complicated motion in real scenes. In this paper, we propose an end-to-end training method \textbf{$\text{A}^2\text{OF}$} for video frame interpolation with event-driven \textbf{A}nisotropic \textbf{
A}djustment of \textbf{O}ptical \textbf{F}lows. Specifically, we use events to generate optical flow distribution masks for the intermediate optical flow, which can model the complicated motion between two frames. Our proposed method outperforms the previous methods in video frame interpolation, taking supervised event-based video interpolation to a higher stage.
\end{abstract}

\section{Introduction}  \label{sec_introduction}

Video frame interpolation(VFI) is a challenging task in computer vision, which is widely used in slow motion video generation, high rate frame conversion and video frames recovery, etc. The goal of VFI is to synthesize nonexistent intermediate frames between two consecutive frames. However, it is hard to synthesize high-quality intermediate frames due to the lack of corresponding motion information.

\begin{figure}
\centering
\includegraphics[width=\linewidth]{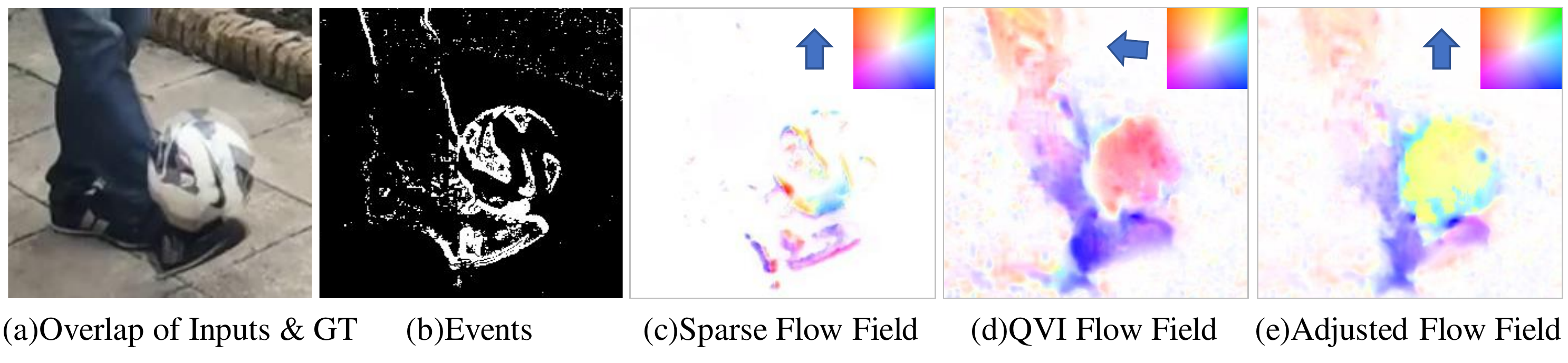}
\caption{The visualization of the generated optical flow. (c) demonstrates the sparse optical flow in Time Lens~\cite{tulyakov_time_2021}. (d) demonstrates the final intermediate optical flow in QVI~\cite{QVI2019Xu}. (e) demonstrates the final intermediate optical flows generated through the proposed model. These optical flows are used to warp the original frames to get initial intermediate frames. The arrows in these pictures denote the directions of football motion.}
\label{fig:optical_example}
\end{figure}

Optical flow is a common tool in VFI. SuperSloMo~\cite{SuperSloMo2018Jiang} linearly represent the intermediate optical flow from the target frame to original frame with bi-directional optical flow. They warp, blend and refine the original frames to obtain target frames. The key assumption in their method is uniform motion along a straight line, which runs counter to the laws of nonlinear motion in real world. Besides, they synthesize the intermediate optical flow with the same coefficient to the different directions in bi-directional optical flows. In this isotropic way, they can not obtain the inaccurate intermediate flow optical which has the different directions with bi-directional optical flow. Aiming at anisotropic adjustment of intermediate flow, some works try to extract more information through pretrained models, such as depth information~\cite{Depth2019Bao} and contextual information~\cite{Context2018Niklaus}. 
More complex motion assumptions are designed in some succeeding works, such as QVI~\cite{QVI2019Xu} and EQVI~\cite{EQVI2020Liu}.  However, it is still difficult for them to describe the actual movement under the absence of intermediate motion information.As shown in Fig.~\ref{fig:optical_example}(d), the intermediate optical flows obtained based on this inaccurate assumption fail to describe the correct direction of football. 

Event streams can record accurate motion information between two consecutive frames, which help to solve the absence of information in VFI model. Time Lens~\cite{tulyakov_time_2021} uses event streams to compensate for motion information. Event streams generated through event camera asynchronously record the motion information for a period of time with the merits of high temporal resolution and low latency. Time Lens directly uses event to synthesize intermediate frame and estimate intermediate optical flow. The pre-warped frames through intermediate optical flow are aligned and refined by event-driven synthesis frames to obtain the final target frames. There are some limitations: although modeling motion is avoided in the process of obtaining intermediate optical flow, the warped result is limited by the sparse flow field estimated through event streams, as shown in Fig.~\ref{fig:optical_example}(c), which lacks the dense features of the RGB images. The predicted sparse optical flow field is also easily affected by the noise of the event itself, which affects the final result. Besides, their model is not an end-to-end training model.

In this paper, we propose a video frame interpolation method \textbf{$\text{A}^2\text{OF}$} with event-driven \textbf{A}nisotropic \textbf{A}djustment of \textbf{O}ptical \textbf{F}low in an end-to-end training manner. Specifically, our model is based on bi-directional optical flows model, such as Super SloMo~\cite{SuperSloMo2018Jiang}. Firstly, we adopt and tailor IFNet in~\cite{RIFE} in order to better model the motion from both event streams and frames. After getting the bi-directional optical flow, instead of linearly weighting them in both horizontal and vertical directions, anisotropic weights are learnt from events to blend and generate intermediate optical flow in the orthogonal directions. As shown in Fig.~\ref{fig:optical_example}(e), the optical flows generated from our proposed models can demonstrate the correct direction of the football. Besides, we design an event-driven motion consistency loss based on the change of intensity to further improve the performance of our proposed model. We carry out a series of experiments to evaluate our proposed model on both synthetic and real event-frame datasets. Experimental results show that our model achieves the-state-of-art performance in all evaluated datasets. 

Our main contributions can be summarized as:

1. To address the limitation in video frame interpolation model based on bi-directional optical flow, we design event-driven optical flow distribution masks generation module to provide anisotropic weights for the different directions of the optical flows. 

2. In order to better compensate the motion information in the event to the VFI model, we design an event-driven motion consistency loss based on the change of intensity. 

3. We design an end-to-end training video frame interpolation model, which outperforms the previous methods on both synthetic and real event-frames datasets.


\section{Related Works} \label{sec_related_works}

\textbf{Video Frame Interpolation.} The goal of VFI is to predict the intermediate frames between the input frames. Some methods~\cite{long2016learning,liu2017video,niklaus2017video} mainly focused on single-frame interpolation, which is commonly ineffective and inflexible. In order to interpolate several frames at any time between consecutive frames,  SuperSloMo~\cite{SuperSloMo2018Jiang} designs a video frame interpolation model based on optical flows. According to the uniform motion assumption, they linearly aggregate bi-directional optical flows to obtain intermediate optical flow from the target frame to the original frame in an isotropic way. The final synthesis intermediate frames are warped and blended from the original frames. Their uniform motion assumption does not well with nonlinear motion in real scenes. Besides, they can not model the real motion which has the different directions with bi-directional optical flow. For anisotropic adjustment of optical flow, some extra information from the pretrained model is added into VFI model, such as depth information~\cite{Depth2019Bao} or contextual information~\cite{Context2018Niklaus}. QVI~\cite{QVI2019Xu} and EQVI~\cite{EQVI2020Liu} design more complex motion assumptions, however, which may still deviates from the way that objects move in the real scenes. The absence of motion information between original frames is the main cause of inaccurate motion estimation and low-quality frames interpolation.  They may have poor frame interpolation ability when encountering complicated motion scenes due to inaccurate motion descriptions. We propose event-driven distribution masks generation module in the bi-directional model to obtain intermediate optical flows in an anisotropic way for the different directions of optical flow. With the help of event streams, we can get more accurate motion descriptions and high-quality interpolation results. 


\textbf{Event Camera.} Through the neuromorphic sensor like dynamic vision sensors (DVS), event camera generates the high-dynamic-range event data under low power consumption~\cite{gallego2019event}. Previous works demonstrate that the event-frame-based models show promising potential in visual tasks like image deblurring ~\cite{deblur2019Pan,jiang2020learning,lin2020learning}, high-dynamic-range image restoration ~\cite{wang2019event,han2020neuromorphic,zhang2020learning}, and VFI ~\cite{VFI2020wang,tulyakov_time_2021,he_timereplayer:_2022}. Some methods~\cite{deblur2019Pan,VFI2020wang,tulyakov_time_2021} avoid modeling the process of motion. The former two methods mix the feature of event and frame through neural networks to directly predict intermediate frames. The disturbance of threshold in event camera has a negative influence on the performance of their models. Time Lens~\cite{tulyakov_time_2021} directly predicts intermediate optical flow through event streams, which limits the accuracy of optical flows estimation due to the absence of visual detail in RGB frames. TimeReplayer~\cite{he_timereplayer:_2022} proposes an unsupervised learning method to aid the video frame interpolation process. Our model is designed based on bi-directional optical flows model. We introduce event streams into our model for the aggregation of bi-directional optical flows in an anisotropic way. The performance of our VFI model is further improved.

\section{Method}        \label{sec_method}

\subsection{Revisiting Bi-directional Optical Flow VFI Model}   \label{subsec_review}
\begin{figure}
\centering
\includegraphics[width=\linewidth]{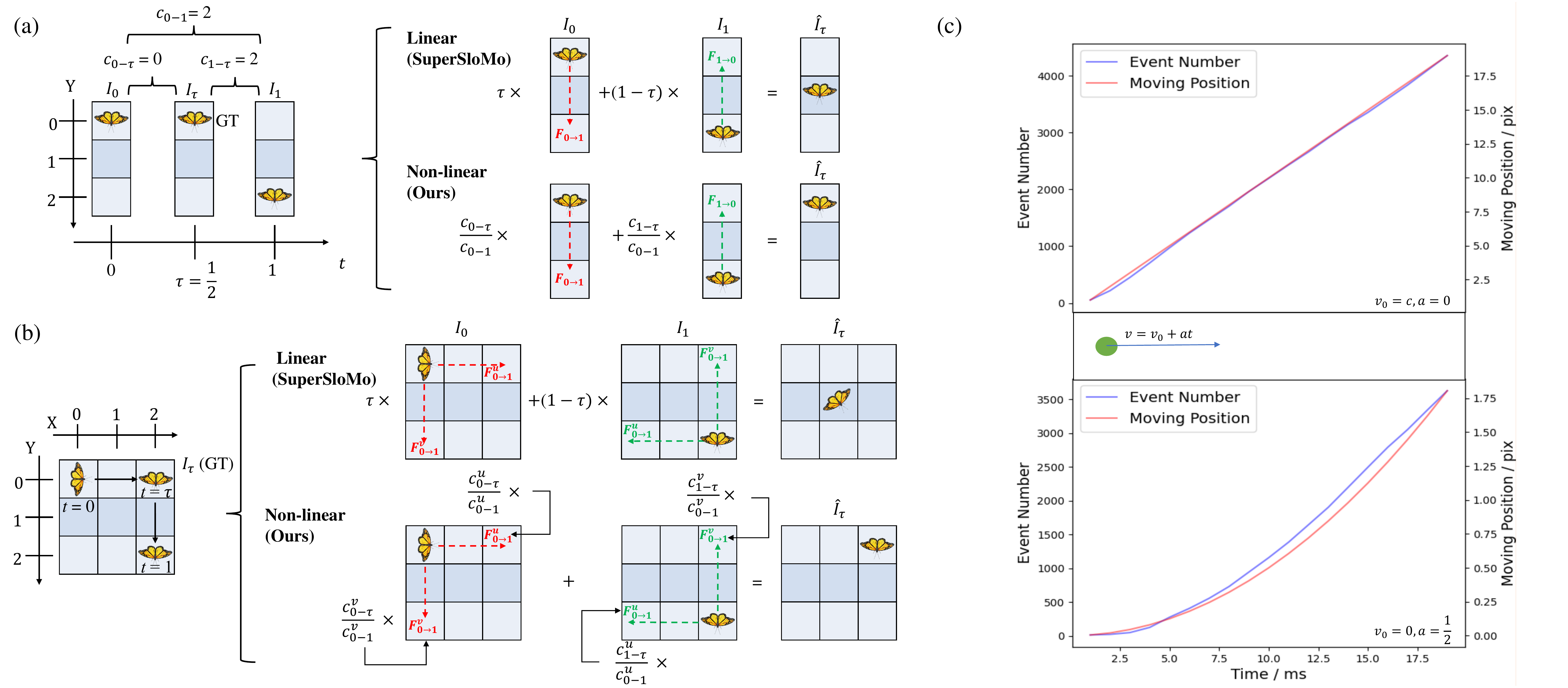}
\caption{(a-b) A toy example for the butterfly moving in $0-1$ period in one-dimensional and two-dimensional space, respectively. For two-dimensional movements, we use the number of events to generate optical flow masks in different directions. (c) Two enlightening examples on the relationship between the motion(the right axis) and event count(the left axis) in uniform rectilinear motion(the top graph) and uniformly accelerated rectilinear motion(the bottom graph). The unit of velocity $v_0$ and $v$ are pixels per time step, and the unit of acceleration $a$ is pixels per time step squared. The time step is 3ms. }
\label{fig:example_for12d}
\end{figure}
Given two consecutive frames $I_0$ and $I_1$ in a video, we can obtain bi-directional optical flow $F_{0\to1}$ and $F_{1\to0}$. $F_{0\to1}$ and $F_{1\to0}$ are the optical flows from $I_1$ to $I_0$ and the optical flow from $I_0$ to $I_1$, respectively. $F_{\tau \to 0}$ and $F_{\tau \to 1}$ are the intermediate optical flows from the target frame to original frames $I_0$ and $I_1$, respectively. Both $F_{\tau \to 0}$ and $F_{\tau \to 1}$ can be represented by bi-directional optical flows $F_{0\to1}$ and $F_{1\to0}$ as shown in Eq.~(\ref{F_t_0}) and Eq.~(\ref{F_t_1}):
\begin{equation} \label{F_t_0}
    F_{\tau \to 0} = g(\tau, F_{1 \to 0}, M_{\text{etra}}) ~or~ g(\tau, F_{0 \to 1}, M_{\text{etra}}),
\end{equation}
\begin{equation} \label{F_t_1}
    F_{\tau \to 1} = g(1 - \tau, F_{1 \to 0}, M_{\text{etra}}) ~or~ g(1 - \tau, F_{0 \to 1}, M_{\text{etra}}).
\end{equation}

We can blend the optical flow obtained from $F_{1 \to 0}$ and $F_{0 \to 1}$ to get more accurate optical flows based on temporal consistency. The final intermediate optical flows can be represented as:
\begin{equation} \label{F_t_0_final}
    F_{\tau \to 0} = \text{Blend}_{F}(g(\tau, F_{1 \to 0}, M_{\text{etra}}), g(\tau, F_{0 \to 1}, M_{\text{etra}}); \theta_{0-\tau}),
\end{equation}
\begin{equation} \label{F_t_1_final}
    F_{\tau \to 1} = \text{Blend}_{F}(g(1 - \tau, F_{1 \to 0}, M_{\text{etra}}), g(1 - \tau, F_{0 \to 1}, M_{\text{etra}}); \theta_{\tau - 1}),
\end{equation}
where $\text{Blend}_{F}(\cdot)$ denotes blending operation between initial intermediate optical flows. $\theta_{\cdot}$ denotes the parameters in the blending operations, such as linearly weight in temporal consistency~\cite{SuperSloMo2018Jiang} or parameters in convolution neural networks. 

According to the final intermediate optical flows, original frames $I_0$ and $I_1$ are warped through bilinear interpolation~\cite{bilinear_function} and refined through convolution neural network to get pre-warped intermediate frames $\hat{I}_{0\to\tau}$ and $\hat{I}_{1\to\tau}$.  The final intermediate frame $\hat{I_{\tau}}$ can be represented as the mixture of $\hat{I}_{0\to\tau}$ , $\hat{I}_{1\to\tau}$ and extra pre-warped information such as visibility maps~\cite{SuperSloMo2018Jiang}, depth information~\cite{Depth2019Bao} and contextual information~\cite{Context2018Niklaus} as shown in Eq.~(\ref{final_result}):
\begin{equation} \label{final_result}
    \hat{I_{\tau}} = \text{Blend}_{I}(\hat{I}_{0\to\tau}, \hat{I}_{1\to\tau}, M_{\text{extra}}),
\end{equation}
where $\text{Blend}_{I}$ denotes blending operations between pre-warped intermediate frames.

\textbf{A Toy Example.} We take the VFI model in SuperSloMo~\cite{SuperSloMo2018Jiang} as an example model as shown in Fig.~\ref{fig:example_for12d}(a) and
Fig.~\ref{fig:example_for12d}(b) for motion in one dimension and two dimensions, respectively. For the convenience of description, we take SuperSloMo as an example model. As shown in Fig.~\ref{fig:example_for12d}(a), a butterfly rests at location $Y_0$ in $0-\tau$ period and flies from location $Y_0$ to $Y_2$ in $\tau - 1$ period. $I_0$ records a butterfly locates in $Y_0$, while $I_1$ records a butterfly locates in $Y_2$. For SuperSloMo model, which assumes the butterfly moves at a constant speed along a line, the synthesized frame $I_{\tau}$ shows the butterfly locates in $Y_1$ when $\tau = \frac{1}{2}$. However, the actual coordinates of the butterfly are at $Y_0$. The frame 
interpolation of the butterfly in two-dimensional motion is similar to that in one-dimensional as shown in Fig.~\ref{fig:example_for12d}(b). The butterfly is predicted at location $Y_{(1, 1)}$, while the actual coordinates of the butterfly are at $Y_{(0, 2)}$. The reason for this phenomenon is that they synthesize the intermediate optical flow in an isotropic way. All directions of optical flow are linearly aggregated with the same coefficients. Note that there is no motion in the vertical direction in $0 - \tau$  period in Fig.~\ref{fig:example_for12d}. So the optical flow in the vertical direction should not be taken into account when we calculate the intermediate optical flow $F_{\tau \to 0}$. However, due to the absence of intermediate motion information between two frames, it is difficult to predict the different coefficients in an anisotropic way. Thus, we introduce the event data that records the real motion information into our model and use the event to estimate the optical flow mask to better model the motion in the corresponding period. 


\subsection{Event-driven Optical Flow Mask}

Let’s come back to the one-dimensional condition in Fig.\ref{fig:example_for12d}(a), no motion is observed in $0-\tau$, leading to no event streams generated. However, the event is generated in $\tau-1$ due to the motion of the butterfly. If we regard every change as an event, the polarities of the event data $E_{0\rightarrow\tau}$ of such the $3\times1$ one-dimensional map could be encoded as $\left[0,\ 0,\ 0\right]^T$ at the timestamp $t=\tau$ in $0 - \tau$ period, and that of $E_{\tau\rightarrow1}$ is $\left[-1,\ 0,\ 1\right]^T$ at the timestamp $t = 1$ in $\tau - 1$ period. The total event data in $0-1$ period could be encoded as $\left[-1,\ 0,\ 1\right]^T$, which is the same as the event data $E_{\tau\rightarrow1}$. In this simple example, we could find that the event data could encode the ground-truth movement, which could be possibly used to distribute the bi-directional optical flows. We carry out a simulation experiment that a pixel of ball moves in uniform rectilinear motion or uniformly accelerated rectilinear motion and draw the curve of moving position and the curve of event number along the time. As shown in Fig.~\ref{fig:example_for12d}(c),
the trend of the moving position curve is basically consistent with the velocity-time curve, and the event count can describe the slowness.

Herein, we propose event-driven optical flow masks $\omega_{\cdot}$ that anisotropically determines the weights of the bi-directional optical flows. The optical flow $F_{0\rightarrow1}$ contributes nothing to the synthesis frame ${\hat{I}}_\tau$ due to no motion and no event occurrence in $0-\tau$ period, while the optical flow $F_{1\rightarrow0}$ is more important to the synthesis frame $I_\tau$ due to more motion and event from the butterfly in $\tau - 1$ period. Thus, the distribution of $F_{0\rightarrow1}$ should be set as $0$ while the distribution of $F_{1\rightarrow0}$ should be set as $1$ to generate the accurate intermediate flows. We find the distribution of bi-directional optical flow can be calculated by the ratio between the number of event occurrences in target period and that in the total period. Based on the above analysis, Eq.~(\ref{F_t_0}) and Eq.~(\ref{F_t_1}) can be rewrited as follows:
\begin{equation} \label{F_t_0_our}
    F_{\tau \to 0} = \omega_{0 -  \tau} \cdot F_{1 \to 0} ~ \text{or} ~ - \omega_{0 - \tau}\cdot  F_{0 \to 1},
\end{equation} 
\begin{equation} \label{F_t_1_our}
    F_{\tau \to 1} = \omega_{1 - \tau}\cdot  F_{0 \to 1} ~ \text{or} ~ - \omega_{1 - \tau}\cdot  F_{1 \to 0},
\end{equation} 
where $\omega_{0-\tau}$ and $\omega_{1-\tau}$ denotes the optical flow mask of bi-directional optical flows in $0-\tau$ period and $\tau - 1$. We can obtain these weight through event streams as shown in Eq.~(\ref{weight_our}):
\begin{equation} \label{weight_our}
    \omega_{0-\tau}=\frac{c_{0-\tau}}{c_{0-1}} ~\text{and}~ \omega_{1-\tau}=\frac{c_{1-\tau}}{c_{0-1}},
\end{equation}
where $c_{\cdot}$ denotes the number of event occurrences in target period. Note that large motion does not necessarily lead to large ratio: the optical flow between two frames represent the total amount of motion, while the ratio Eq.~(\ref{weight_our}) indicates how the motion distributes over time.

\begin{figure}[t]
\centering
\includegraphics[width=\linewidth]{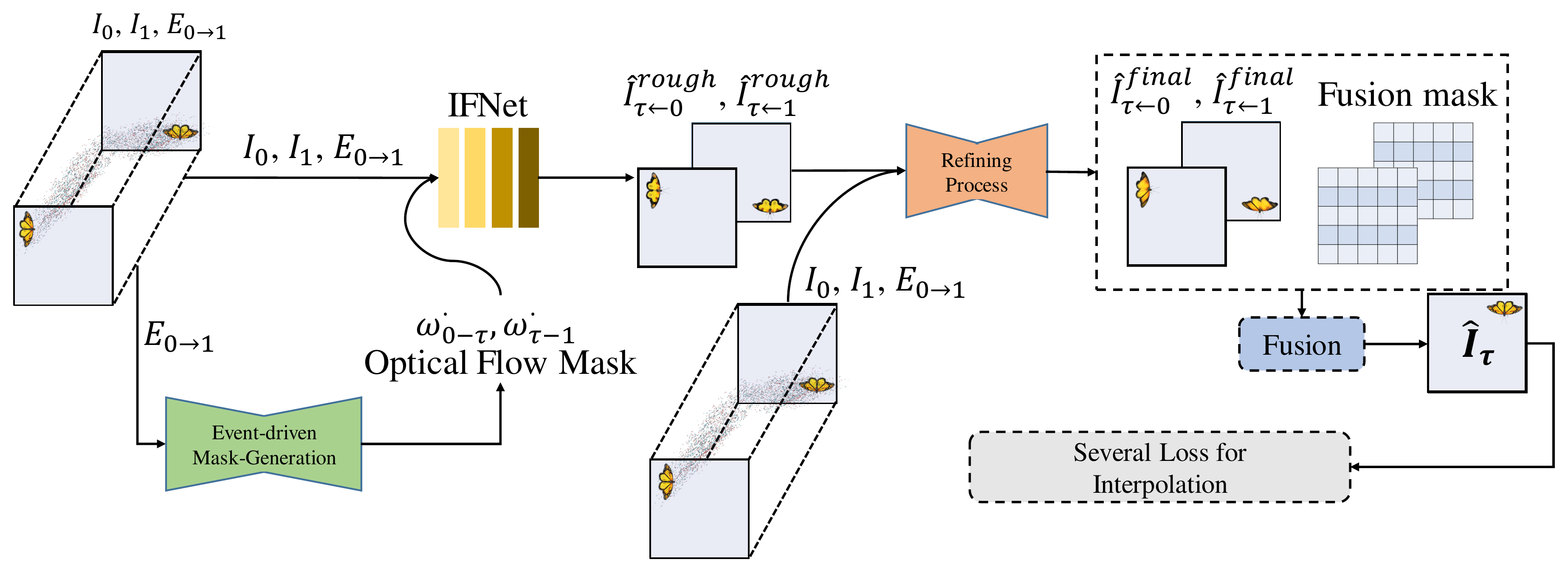}
\caption{The pipeline of our proposed method \textbf{$\text{A}^2\text{OF}$}. The event is input to Event-driven Mask Generation network to get optical flow masks $\omega_{0-\tau}$ and $\omega_{\tau-1}$. We modify the structure of the IFNet~\cite{RIFE} so that it can input event and use anisotropic flow mask as shown in supporting information. Two consecutive frames, corresponding event and optical flow mask are input to tailored IFNet to get intermediate optical flows and warp the original frames. The pre-warped frames $\hat{I}_{\tau\to0}^{\text{rough}}$ and $\hat{I}_{\tau\to1}^{\text{rough}}$ are fused to final intermediate frame $\hat{I}_{\tau}$under the supervision of a series of loss.} 
\label{fig:pipeline}
\end{figure}
In order to better analyze the motion in two-dimensional space and model the motion between two frames, we calculate different weights in the orthogonal directions as shown in Fig.~\ref{fig:example_for12d}(b). Specifically, as shown in Fig.~\ref{fig:example_for12d}(b) and Fig.~\ref{fig:example_for12d}(c), the original optical flow $F_{1\to0}$ can be divided into $F_{1 \to 0}^{u}$ in the horizontal direction and $F_{1 \to 0}^{v}$ in the vertical direction. $F_{0\to1}$ can be decomposed in the same way as $F_{0 \to 1}^{u}$ and $F_{0 \to 1}^{v}$. The subscripts of these decomposed optical flows indicate the optical flows are in which period, and the superscripts indicate the optical flows are in which direction. Note that we can still obtain intermediate optical flows in the different directions based on Eq.~(\ref{F_t_0_our}-\ref{weight_our}). Based on the time consistency, we can further combine initial intermediate optical flows $F_{\tau\to0}^{\cdot}$ or $F_{\tau\to1}^{\cdot}$ from $F_{1\to0}^{\cdot}$ and $F_{0\to1}^{\cdot}$, where $\cdot$ in the right corner denotes the direction of optical flow, such as horizontal or vertical direction. The final intermediate optical flows can be represented as:
\begin{equation} \label{F_t_0_u}
    F_{\tau\to0}^{u} = - (1 - \tau) \cdot \omega_{0-\tau}^{u} \cdot  F_{0\to1}^{u} + \tau \cdot \omega_{0-\tau}^{u} \cdot F_{1\to0}^{u},
\end{equation}
\begin{equation} \label{F_t_1_u}
    F_{\tau\to1}^{u} = (1 - \tau) \cdot \omega_{1-\tau}^{u} \cdot  F_{0\to1}^{u} - \tau \cdot \omega_{1-\tau}^{u} \cdot F_{1\to0}^{u}.
\end{equation}
We can also obtain intermediate optical flow in the vertical direction as shown in Eq.~(\ref{F_t_0_u}-\ref{F_t_1_u}).

Note that the event contains a certain amount of noise due to system noise and unfixed thresholds. Calculation to the weight with unprocessed events will result in inaccurate weight prediction. Thus, we propose a U-Net style convolution neural network named Event-driven Mask Generation Network to process the input event data. The architecture of Event-driven Mask Generation Network can be found in Supporting Information. As shown in Fig.~\ref{fig:pipeline}, our event-driven mask generation network processes the input event $E_{0 \to 1}$ and outputs two channels feature map which encodes the number of events at each pixel in $0-\tau$ or $\tau-1$ period. The final weight $\omega_{0 \to \tau}^{\cdot}$ and $\omega_{1 \to \tau}^{\cdot}$ is calculated according to Eq.~(\ref{weight_our}) with these two-channel feature map in the horizontal and vertical direction. The spatial size of $\omega_{0 \to \tau}^{\cdot}$ and $\omega_{1 \to \tau}^{\cdot}$ is with the same spatial size $H \times W$ as bi-directional optical flows. These event-driven optical flow masks are used to obtain the nonlinear intermediate flows $F_{\tau\rightarrow0}^{\cdot}$ and $F_{\tau\rightarrow1}^{\cdot}$ as shown in Eq.~(\ref{F_t_0_u}-\ref{F_t_1_u}).

\subsection{Pipeline for Our Event-driven Video Interpolation Model}
In this section, we will specifically show the representation of events and the pipeline of our event-driven video interpolation model. 

\subsubsection{Event Representation.}

An original event $e$ can be represented as a four-element tuple $(x_e, y_e, p_e, t_e)$, where $x_e$ and $y_e$ denote spatial coordinates. $p_e$ denotes the polarity of the event and $t_e$ denotes the time of occurrence for the event. We should convert those event streams into 2-D frames to input the convolution neural network. Our event streams during the target period are represented as a four-channel frame as shown in~\cite{event_representation}. The first and second channels encode the number of positive and negative polarities of events at each pixel, respectively. The third and fourth channels encode the timestamps of the latest triggered positive and negative events, respectively. The represented event data between two consecutive frames $I_0$ and $I_1$ is set as $E_{0\to1}$.

\subsubsection{Event-driven Optical Flows Estimation.} 

The image frames encode the details of the motion with the low temporal resolution, while the event records the motion information with high temporal resolution. A natural idea is to use these two complementary data together to predict the bi-directional optical flow. Time Lens~\cite{tulyakov_time_2021} uses events to directly predict intermediate optical flow which lacks the details of the motion information from frames.

Thus, we take the frame data $I_0$, $I_1$, event $E_{0\rightarrow1}$ and anisotropic optical flow weights as the input to IFNet~\cite{RIFE} to estimate the bi-directional optical flow and warp the original frames as shown in Fig.~\ref{fig:pipeline} and Supporting Information. We carefully tailor the IFNet and make it suitable for our task. Firstly, our tailored IFNet can process both event and image frames. In this way, the synthetic intermediate optical flow can correctly model the target motion due to the complementary of the two types of data with our anisotropic optical flow weights according to Eq.~(\ref{F_t_0_u}-\ref{F_t_1_u}). Then, we deeper the IFNet to better process these two types of data. Our tailored IFNet has 4 blocks. Next, the optical flow mask from Event-driven Mask Generation network is input to each block in our tailored IFNet for blending and synthesizing intermediate optical flow. The output of IFNet is rough interpolated frames $\hat{I}_{\tau \gets 1}^{\text{rough}}$ and $\hat{I}_{\tau \gets 0}^{\text{rough}}$ which will be input to refinement network.

\subsubsection{Refinement Network.}

The warped images ${\hat{I}}_{\tau\gets0}^{\text{rough}}$, ${\hat{I}}_{\tau\gets1}^{\text{rough}}$ could encode the most nonlinear motion but poorly around the stationary objects due to the absence of event data in the static regions. To make the intermediate frames better, a refining process is necessary like the previous work ~\cite{SuperSloMo2018Jiang,QVI2019Xu,tulyakov_time_2021}. Except for the additional input event, the overall structure of our refinement network is similar to that in~\cite{SuperSloMo2018Jiang}. Our refinement network is a U-Net style network with 6 encoders and 5 decoders with a shortcut between encoder and decoder of the same spatial scale. The details of our refinement network are provided in supporting information.As shown in Fig.~\ref{fig:pipeline}, we input the data $I_0$, $I_1$, $E_{0\rightarrow1}$,${\hat{I}}_{\tau\gets0}^{\text{rough}}$, ${\hat{I}}_{\tau\gets1}^{\text{rough}}$ and bi-directional optical flow $F_{0\to1}$, $F_{1\to0}$ into a sub-network.  Herein, the output is two fusion maps $V_{\tau\gets0}$, $V_{\tau\gets1}$ and two refined frames ${\hat{I}}_{\tau\gets0}^{\text{final}}$ and ${\hat{I}}_{\tau\gets1}^{\text{final}}$.Thus, the final intermediate frames ${\hat{I}}_\tau$ is defined after the Fusion process of $\text{Fusion}(\cdot)$:
\begin{equation}
  \begin{aligned}
  \hat{I}_{\tau} &=\operatorname{Fusion}\left(I_{\tau \leftarrow 0}^{\text {final }}, I_{\tau \leftarrow 1}^{\text {final }}, V_{\tau \leftarrow 0}, V_{\tau \leftarrow 1}\right) \\
  &=V_{\tau \leftarrow 0} \cdot I_{\tau \leftarrow 0}^{\text {final }}+V_{\tau \leftarrow 1} \cdot I_{\tau \leftarrow 1}^{\text {final }},
  \end{aligned}
  \end{equation}
where $V_{\tau \leftarrow 0}$ and $V_{\tau \leftarrow 1}$ are two visibility maps which encode whether the objects are occluded. The pixel-addition of two visibility map equals 1, following~\cite{SuperSloMo2018Jiang}.

\subsubsection{Loss Function.}

Our event-driven video frame interpolation model can be trained in an end-to-end manner under the combination of event-driven motion consistency loss $\mathcal{L}_{\text{MC}}$, the reconstruction loss $\mathcal{L}_{\text{rec}}$, the perceptual loss $\mathcal{L}_{\text{per}}$, the warped loss $\mathcal{L}_{\text{warp}}$ and the smoothness loss $\mathcal{L}_{\text{smooth}}$. Note that the reconstruction loss $\mathcal{L}_{\text{rec}}$, the perceptual loss $\mathcal{L}_{\text{per}}$,the warped loss $\mathcal{L}_{\text{warp}}$ and the smoothness loss $\mathcal{L}_{\text{smooth}}$ are similar to~\cite{SuperSloMo2018Jiang}. 

Event-driven motion consistency loss $\mathcal{L}_{\text{mc}}$ measures the gap between estimated event count map $\hat{E}_{\cdot}^{\text{count}}$ and real event count map $E_{\cdot}^{\text{count}}$. We take event count map $\hat{E}_{0\to\tau}^{\text{count}}$ as an example. $\hat{E}_{0\to\tau}^{\text{count}}$ is calculated with the difference between two frames $I_{\text{diff}}$ which is represented as
\begin{equation} \label{I_diff}
    I_{\text{diff}} = \text{log}(\frac{I_{\tau \gets 0}^{\text{final}}}{I_0}).
\end{equation}
Due to unknown and inflexible thresholds in event camera, it is difficult to directly get event count map through Eq.~(\ref{I_diff}). So, we binarize the event count to reflect whether the event appears or not in each pixel. Specifically, in the event count map, 1 denotes there is at least one event while 0 denotes there is no event. How to binarize the $I_{\text{diff}}$ to obtain estimated event count map $\hat{E}_{0\to\tau}^{\text{count}}$? Firstly, we get $E_{0\to\tau}^{\text{count}}$ through the first and second channel in $E_{0\to\tau}$:
\begin{equation}
    E_{0\to\tau}^{\text{count}} = \text{sgn}(E_{0\to\tau}[0:2]),
\end{equation}
where $\text{sgn}(\cdot)$ denotes the sign function and $[0:2]$ denotes the first and second channel of a tensor. Then, we sum all elements in $E_{0\to\tau}^{\text{count}}$ along each channel to obtain binarized threshold $t_{\text{positive}}$ and $t_{\text{negative}}$, respectively. These thresholds record the number of locations where the event has occurred. Next, we sort $I_{\text{diff}}$ in descending order and select the top $t_{\text{positive}}$ value $N_{t_{\text{positive}}}$ and the bottom $N_{t_{\text{negative}}}$ value as binarized thresholds for $I_{\text{diff}}$.
\begin{equation}
    \hat{E}_{0\to\tau}^{\text{count}} = [\text{sgn}(I_{\text{diff}} - N_{t_{\text{positive}}}), \text{sgn}(N_{t_{\text{negative}}} - I_{\text{diff}})],
\end{equation}
where $[\cdot, \cdot]$ denotes concat function in tensor. $\text{sgn}(I_{\text{diff}} - N_{t_{\text{positive}}})$ represents where there may be positive events. We want $\hat{E}_{0\to\tau}^{\text{count}}$ and $E_{\cdot}^{\text{count}}$ to be exactly the same, which means that the motion information recorded by both is almost the same. The definition of $\mathcal{L}_{\text{mc}}$ is
\begin{equation}
    \mathcal{L}_{\text{mc}} = \frac{1}{N}||{G}(\hat{E}_{0\to\tau}^{\text{count}}) - {G}(E_{0\to\tau}^{\text{count}})||_1 + \frac{1}{N}||{G}(\hat{E}_{\tau\to1}^{\text{count}}) - {G}(E_{\tau\to1}^{\text{count}})||_1 ,
\end{equation}
where $G(\cdot)$ denotes Gaussian Blur Function. We smooth the $\hat{E}_{\cdot}^{\text{count}}$ with $G(\cdot)$ to alleviate the effect of noise in the event. The definition of the reconstruction loss, the perception loss, the warping loss and the smoothness loss here are the same as SuperSloMo~\cite{SuperSloMo2018Jiang} does.



The total loss $\mathcal{L}$ of our model is 
\begin{equation} \label{total_loss}
    \mathcal{L} = \lambda_{mc}\cdot\mathcal{L}_{\text{mc}} + \lambda_{rec}\cdot\mathcal{L}_{\text{rec}} + \lambda_{per}\cdot\mathcal{L}_{\text{per}} + \lambda_{warp}\cdot\mathcal{L}_{\text{warp}} + \lambda_{smooth}\cdot\mathcal{L}_{\text{smooth}}.
\end{equation}
Note that all the loss weights in Eq.~(\ref{total_loss}) is set empirically on the validation set. Specifically, $\lambda_{mc} = 1.0$, $\lambda_{rec} = 1.0$, $\lambda_{per} = 0.2$, $\lambda_{warp} = 0.8$ and $\lambda_{smooth} = 0.8$.

\section{Experiments}   \label{sec_experiments}

\subsection{Implementation Details}
Our proposed model and all experiments are implemented in Pytorch~\cite{paszke2019pytorch}. We use adam optimizer with standard settings in~\cite{kingma2014adam}. For training, our model is trained end-to-end on 4 NVIDIA Tesla V100 GPUs for the total 500 epochs. The batch size of each training step is 28. The initial learning rate is $10^{-4}$ and is 
multiplied by 0.1 per 200 epochs. We calculate the peak-signal-to-noise ratio (PSNR), structural similarity(SSIM), and interpolation-error (IE) as the quantitative metric to evaluate the performance of our proposed method.

\subsection{Datasets}
We firstly evaluate our model on three common high-speed synthetic datasets with synthetic events: Adobe240~\cite{adobe}, GoPro~\cite{GoPro} and Middlebury~\cite{Middlebury}. Besides, we also evaluate our model on the real frame-event dataset: High-Quality Frames (HQF)~\cite{HQF}, High Speed Event-RGB(HS-ERGB)~\cite{tulyakov_time_2021}. All the synthetic training sets are collected to train our model. Then, we compare with other previous state-of-the-art methods.

\subsubsection{Synthetic datasets.} Adobe240 consists of 112 different sequences for training and 8 sequences for testing. GoPro dataset consists of 22 different videos for training and 11 videos for testing. Captured by GoPro cameras, both of them own 240 fps and $1280\times720$ resolutions in all the sequences. We use ESIM~\cite{rebecq2018esim} to generate event streams between two consecutive frames. We crop patches with spatial size $384 \times 384$ for training.

\begin{figure*}[t] 
    \centering 
    \includegraphics[width=1\textwidth]{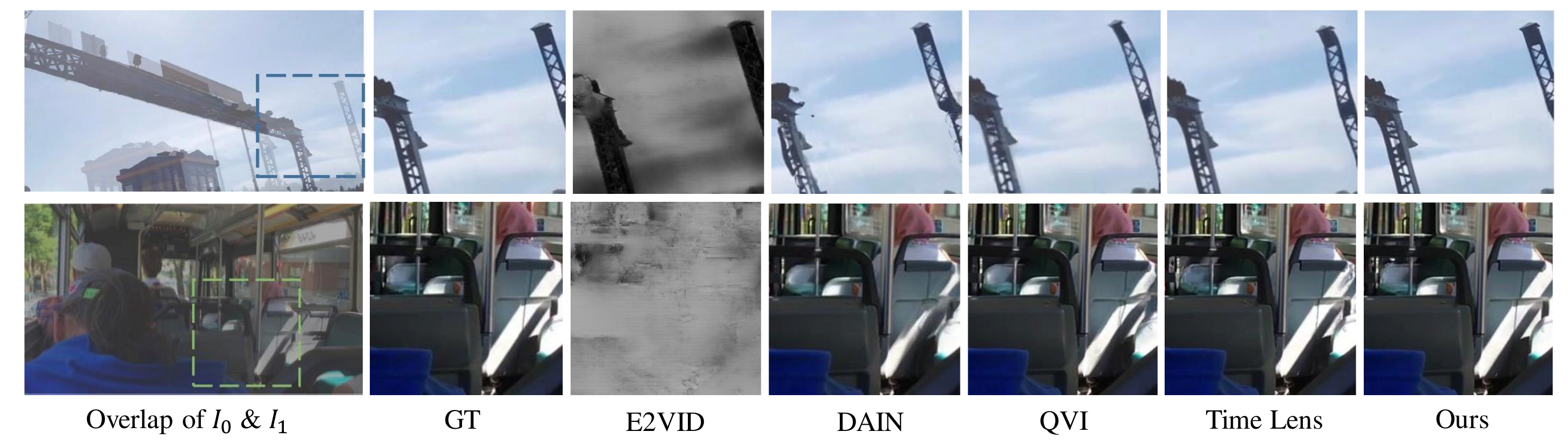}
    \caption{Visual comparisons with different methods on the synthetic dataset Adobe240.} 
    \label{Fig:show_on_synthetic_data} 
  \end{figure*}

\begin{table}[t]
\renewcommand\arraystretch{1}
\centering
\vspace{-0.1cm}
\caption{Quantitative comparison with previous methods on the synthetic datasets Adobe240, GoPro and Middlebury. Note that we directly evaluate the released model in each method without re-training or fine-tuning.}
\resizebox{1\columnwidth}{!}{
\begin{tabular*}{\linewidth}{@{\extracolsep{\fill}}lcccc|ccc}
\hline
\hline
\multicolumn{2}{l}{Adobe240} &\multicolumn{3}{c}{All frames in 7 skips} & \multicolumn{3}{l}{Middle frame in 7 skips} \\
 \hline
 Method & Input & PSNR & SSIM & IE      & PSNR & SSIM & IE \\
\hline
E2VID~\cite{rebecq2019high} &Event     &10.40 &0.570 &75.21    &10.32 &0.573 &76.01   \\
SepConv&RGB &32.31 &0.930 &7.59     &31.07 &0.912 &8.78    \\
DAIN&RGB    &32.08 &0.928 &7.51     &30.31 &0.908 &8.94    \\
SuperSloMo&RGB &31.05 &0.921 &8.19  &29.49 &0.900 &9.68    \\
QVI& RGB    &32.87 &0.939 &6.93     &31.89 &0.925 &7.57    \\
Time Lens&RGB+E&35.47 &0.954 &5.92  &34.83 &0.949 &6.53    \\
Ours($\text{A}^2\text{OF}$)&RGB+E  &\textbf{36.59} &\textbf{0.960} &\textbf{5.58}                   &\textbf{36.21} &\textbf{0.957} &\textbf{5.96} \\ 
\hline

\multicolumn{2}{l}{GoPro} &\multicolumn{3}{c}{All frames in 7 skips} & \multicolumn{3}{l}{Middle frame in 7 skips} \\
\hline
 Method & Input & PSNR & SSIM & IE      & PSNR & SSIM & IE \\
\hline
E2VID&Event     &9.74  &0.549 &79.49    &9.88  &0.569  &80.08   \\
SepConv&RGB &29.81 &0.913 &8.87     &28.12 &0.887  &10.78     \\
DAIN&RGB    &30.92 &0.901 & 8.60    &28.82 &0.863  &10.71    \\
SuperSloMo&RGB &29.54 &0.880 &9.36  &27.63 &0.840  &11.47    \\
QVI& RGB    &31.39 &0.931 &7.09     &29.84 &0.911  &8.57    \\
Time Lens&RGB+E&34.81 &0.959 &5.19  &34.45 &0.951  &5.42    \\
Ours($\text{A}^2\text{OF}$)&RGB+E  &\textbf{36.61} &\textbf{0.971} &\textbf{4.23}                   &\textbf{35.95} &\textbf{0.967} &\textbf{4.62} \\ 
\hline
\multicolumn{2}{l}{Middlebury (other)} &\multicolumn{3}{c}{All frames in 3 skips} & \multicolumn{3}{l}{Middle frame in 3 skips} \\
\hline
 Method & Input & PSNR & SSIM & IE      & PSNR & SSIM & IE \\
\hline
E2VID&Event  & 11.26 & 0.427 & 69.73& 11.12 & 0.407 & 70.35   \\
SepConv&RGB & 25.51 & 0.824& 6.74 & 25.12 & 0.811 & 7.06    \\
DAIN&RGB    & 26.67 & 0.838& 6.17& 25.96& 0.793 & 6.54    \\
SuperSloMo&RGB & 26.14& 0.825& 6.33 & 25.53 & 0.805 & 6.85    \\
QVI& RGB    & 26.31 & 0.827 & 6.58 & 25.72 & 0.798 & 6.73    \\
Time Lens&RGB+E& 32.13 & 0.908& 4.07& 31.57 & 0.893& 4.62    \\
Ours($\text{A}^2\text{OF}$)&RGB+E  &\textbf{32.59} &\textbf{0.916} &\textbf{3.92}                   &\textbf{31.81} &\textbf{0.903} &\textbf{4.13} \\ 
\hline
\hline
\end{tabular*}}
\vspace{-0.3cm}
\label{Tab:performance_on_synthetic_data}
\end{table}


\subsubsection{Real dataset.} High-Quality Frames(HQF) is collected through the DAVIS240 event camera ~\cite{stoffregen2020reducing} which can generate both event streams and corresponding frames. It consists of 14 different frames sequences with the corresponding event streams. The resolutions are $240 \times 180$ in all sequences. We crop patches with spatial size $128 \times 128$ for training. HS-ERGB is collected through Gen4M 720p event camera and FLIP BackFly S RGB camera. This dataset is divided into far-away sequences and close planar scenes. 
\subsection{Comparisons with Previous Methods}
\subsubsection{Synthetic datasets.} We compare our proposed video frame interpolation (VFI) model with previous VFI models. Previous methods can be classified into three categories: frame-based approach, event-based approach and frame-event-based approach. For frame-based approach, we compare our model with SuperSloMo ~\cite{SuperSloMo2018Jiang}, DAIN~\cite{Depth2019Bao}, SepConv~\cite{niklaus2017video}, QVI~\cite{QVI2019Xu}. For the event-based approach, we compare our model with event-based video reconstruction method E2VID ~\cite{rebecq2019high}. For event-frame-based approach, we compare our model with Time Lens ~\cite{tulyakov_time_2021}.  We make a fair comparison on three synthetic datasets Adobe240 ~\cite{adobe}, GoPro ~\cite{GoPro} and Middlebury~\cite{Middlebury}. The comparison results are as shown in Tab.~\ref{Tab:performance_on_synthetic_data}. For training on Adobe240 and GoPro, we select 1 frame from every 8 frames in the original sequence, use the selected frames to form the input sequences, and use the remaining frames as the label for interpolated frames. 
Middlebury is only used for the test due to the sequence length.  Besides, the event streams are generated through ESIM~\cite{rebecq2018esim} between selected frames is also available to our model. For the input frames sequences from selected frames and events, the skipped frames are reconstructed through our proposed method and compared with the ground truth skipped frames. The average performance of the whole 7 skipped frames and the center one are both calculated for fair comparison. Meanwhile, only 3 frames are skipped and used to calculate reconstruction metrics due to the sequence length limitation in Middlebury. The results are summarized in Tab.~\ref{Tab:performance_on_synthetic_data}. Note that we directly evaluate the released model in each comparison method without re-training or fine-tuning.

From Tab.~\ref{Tab:performance_on_synthetic_data}, our proposed method outperforms all the previous methods and achieves the-state-of-art performance on Adobe240, GoPro and Middlebury. The frame-event based approaches outperform the frame-only or event-only approaches. As shown in Fig.~\ref{Fig:show_on_synthetic_data}, due to better use of events with event-driven optical flow mask and event-driven motion consistency loss, our model achieves the best visual quality. Specifically, for the second row in Fig.~\ref{Fig:show_on_synthetic_data}, all objects are static while only the intensity changes. It's a common issue in warping based interpolation method, which is also mentioned by Time Lens. The results indicate improvement by event-driven motion consistency loss in our method.


\begin{table}[t]
\centering
\caption{Quantitative comparison with previous methods on the real frame-event datasets HQF. Note that we directly evaluate the released model in each comparison method without re-training or fine-tuning. }
\begin{tabular*}{\linewidth}{@{\extracolsep{\fill}}lccc|cc}
\hline\hline
\multicolumn{2}{l}{HQF} &\multicolumn{2}{c}{3 skips} & \multicolumn{2}{c}{1 skip} \\
 \hline Method&Input& PSNR   & SSIM   & PSNR & SSIM  
 \\ \hline
E2VID   &Event  &6.70   &0.315   &6.70  &0.315  \\
RRIN    &RGB    &26.11  &0.778   &29.76 &0.874  \\
BMBC    &RGB    &26.32  &0.781   &29.96 &0.875  \\
DAIN    &RGB    &26.10  &0.782   &29.82 &0.875  \\
SuperSloMo &RGB &25.54  &0.761   &28.76 &0.861  \\ 
Time Lens  &RGB+E&30.57 &0.900   &32.49 &0.927  \\
Ours($\text{A}^2\text{OF}$)    &RGB+E  &\textbf{31.85} &\textbf{0.932} 
                &\textbf{33.94} &\textbf{0.945}  \\
\hline
\multicolumn{2}{l}{HS-ERGB (far)} &\multicolumn{2}{c}{7 skips} & \multicolumn{2}{c}{5 skip}  \\
\hline Method&Input& PSNR   & SSIM   & PSNR & SSIM  
 \\ \hline
E2VID   &Event  & 7.01  &0.372   & 7.05 &0.374  \\
RRIN    &RGB    & 23.73 & 0.703 & 25.26 & 0.738  \\
BMBC    &RGB    & 24.14 & 0.710 & 25.62 & 0.742  \\
DAIN    &RGB    & 27.13 & 0.748  & 27.92& 0.780  \\
SuperSloMo &RGB & 24.16 & 0.692  & 25.66 & 0.727  \\ 
Time Lens  &RGB+E& 32.31 & 0.869 & 33.13 & 0.877  \\
Ours($\text{A}^2\text{OF}$)    &RGB+E  &\textbf{33.15} &\textbf{0.883} 
                &\textbf{33.64} &\textbf{0.891}   \\
\hline
\multicolumn{2}{l}{HS-ERGB (close)} &\multicolumn{2}{c}{7 skips} & \multicolumn{2}{c}{5 skip}  \\
\hline Method&Input& PSNR   & SSIM   & PSNR & SSIM  
 \\
 \hline
E2VID   &Event  & 7.68 & 0.427   & 7.73 &0.432  \\
RRIN    &RGB    & 27.46 & 0.800  & 28.69 & 0.813   \\
BMBC    &RGB    & 27.99 & 0.808  & 29.22 & 0.820  \\
DAIN    &RGB    & 28.50 & 0.801  & 29.03 & 0.807  \\
SuperSloMo &RGB & 27.27 & 0.775  & 28.35 & 0.788   \\ 
Time Lens  &RGB+E& 31.68& 0.835  & 32.19 & 0.839  \\
Ours($\text{A}^2\text{OF}$)    &RGB+E  &\textbf{32.55} &\textbf{0.852} 
                &\textbf{33.21} &\textbf{0.865}   \\
\hline\hline
\end{tabular*}
\vspace{-0.5cm}
\label{Tab:performance_on_realfe_data}
\end{table}

\begin{figure*}[t] 
    \centering 
    \includegraphics[width=1\textwidth]{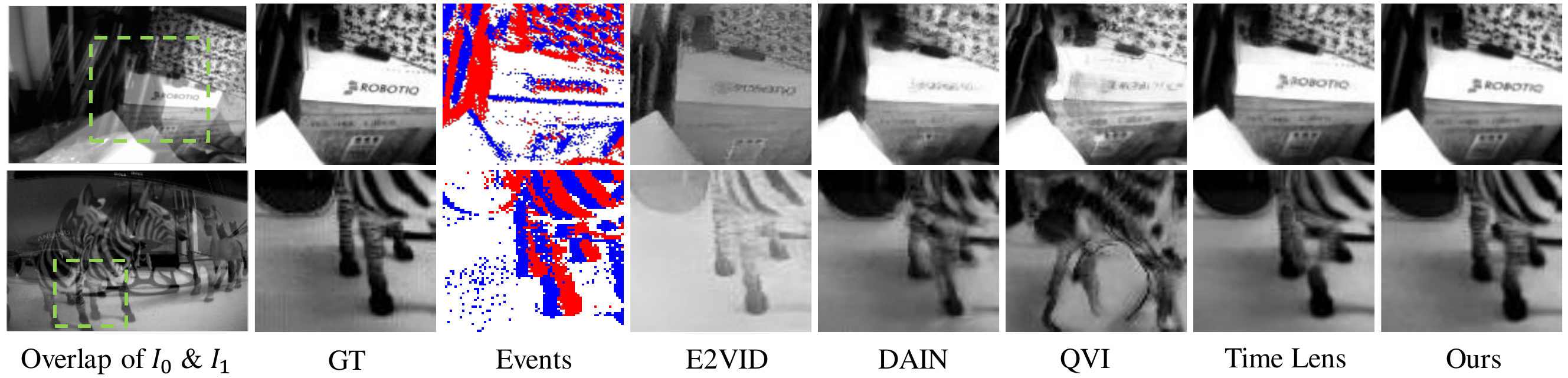}
    \vspace{-0.8cm}
    \caption{Visual comparisons with different methods on HQF with real events.} 
    \label{Fig:show_on_real_data} 
  \end{figure*}

\subsubsection{Real frame-event dataset.} We also evaluate and compare our method with the state-of-the-art methods on the real dataset of HQF: E2VID~\cite{rebecq2019high}, RRIN~\cite{li2020video}, BMBC~\cite{park2020bmbc}, DAIN ~\cite{Depth2019Bao}, SuperSloMo~\cite{SuperSloMo2018Jiang} and Time Lens~\cite{tulyakov_time_2021}. The results are summarized in Tab.~\ref{Tab:performance_on_realfe_data}. We have two experiment settings for training on HQF: Firstly, we select 1 frame from every 4 frames in the original sequences which is viewed as '3-skip' as shown in Tab.~\ref{Tab:performance_on_realfe_data}. Second, we select 1 frame from every 2 frames in the original sequences which is viewed as  '1-skip' as shown in Tab.~\ref{Tab:performance_on_realfe_data}. For HS-ERGB, we select 1 frame from every 8 frames or select 1 frame from evey 6 frames. Note that our model is finetuned with real frames-event datasets based on the model trained on synthetic datasets.

Similar results as the experiments on synthetic datasets could be indicted by quantitative comparison as shown in Tab.~\ref{Tab:performance_on_realfe_data}. As shown in Fig.~\ref{Fig:show_on_real_data}, only our method could reconstruct the legs of the moving horse in the second row.  Due to the absence of visual details from frames in E2VID, synthesized frames through the only event has a large the gap with real intermediate frames. DAIN and QVI fail to synthesize high-quality interpolated frames in the regions with complex motions such as the letters in the first row and the legs in the second row. Their methods cannot model the complicated motion in the real scenes very well.

\subsection{Ablation Studies}
To study the contribution of each module in our proposed model, we perform the ablation study on the GoPro dataset as shown in Tab.~\ref{tab:ablation_study}. Firstly, to evaluate the influence of Event-driven Optical Flow mask, we replace the mask with linear weights in SuperSloMo~\cite{SuperSloMo2018Jiang}.  Experimental results show that our proposed event-driven optical flow mask has a positive effect in VFI model with the improvement of PSNR, SSIM and IE. As for our proposed event-driven motion consistency loss, we train a model without the supervision of this loss. This loss can be used to constrain the motion information coming from the change of intensity and event streams to be similar. Results show that this loss can further improve the performance of our proposed model.

\begin{table}[t]
\centering
\caption{Ablation studies. EDOF denotes our proposed event-driven optical flow mask. MCL denotes our proposed event-driven Motion Consistency Loss.}
\begin{tabular*}{\linewidth}{@{\extracolsep{\fill}}lc|ccc}
\hline\hline 
EDOF & MCL      &PSNR   &SSIM   &IE         \\ \hline

$\times$        & $\checkmark$      &33.82  &0.952  &7.13       \\
$\checkmark$    & $\times$          &35.27  &0.963  &5.75       \\
$\checkmark$    & $\checkmark$      &36.61  &0.971  &4.23       \\
\hline\hline
\end{tabular*}
\vspace{-0.5cm}
\label{tab:ablation_study}
\end{table}

\section{Conclusion}    \label{sec_conclusion}

In this paper, we propose a video frame interpolation method with event-driven anisotropic flow adjustment in an end-to-end training strategy. Besides, we design an event-driven motion consistency loss based on the change of intensity to constrain the gap between the estimated motion information and that from event streams. Instead of proposing complex motion assumptions like previous work, leading to isotropic intermediate flow generation or anisotropic adjustment through learned higher-order motion information, we use events to generate event-driven optical masks for the different directions, assigning the weight of bi-directional optical in intermediate optical flow in an anisotropic way.The proposed event-driven motion consistency loss further improves our method. The experiment results show that our model performs better than previous methods and achieves the state-of-the-art performance.

\clearpage
%
%
\bibliographystyle{splncs04}
\bibliography{egbib}
\pagestyle{headings}
\mainmatter
\def\ECCVSubNumber{4754}  

\title{Video Interpolation by Event-driven\\ Anisotropic Adjustment of Optical Flow} 


\titlerunning{Video Interpolation by Event-driven Anisotropic Adjustment of Optical Flow}
%
\author{Song Wu\inst{\star 1}               \and
Kaichao You\thanks{Song Wu and Kaichao You contribute equally to this paper. Work done while Song Wu, Kaichao You, Yang Tian are interns at Huawei.}\inst{2}                 \and
Weihua He\thanks{Weihua He and Ziyang Zhang are corresponding authors.}\inst{2} \and 
Chen Yang\inst{1} \and Yang Tian\inst{2}    \and 
\\Yaoyuan Wang\inst{1}                      \and 
Ziyang Zhang\inst{\star \star 1}            \and 
Jianxing Liao\inst{1}}

\authorrunning{Wu et al.}
\institute{
Advanced Computing and Storage Lab, Huawei Technologies Co. Ltd 
\\ \email{\{wusong5533, yangchen2017.pku\}@gmail.com}
\\
\email{\{wangyaoyuan1,zhangziyang11,liaojianxing\}@huawei.com} \and 
Tsinghua University
\\ \email{\{ykc20,hwh20,tiany20\}@mails.tsinghua.edu.cn} 
}

\maketitle
\begin{abstract}
This document is our supporting information of our paper \textbf{Video Interpolation by Event-driven Anisotropic Adjustment of Optical Flow.} In this document, we represent our architecture of Event-driven Optical Flow Mask Generation Network and refinement network. Besides, we provide a video demo to show the performance of our proposed methods.
\end{abstract}

\section{The Architecture of Proposed Modules}
\subsection{Event-driven Optical Flow Mask Generation Network.}
\begin{figure}
\centering
\includegraphics[width=\linewidth]{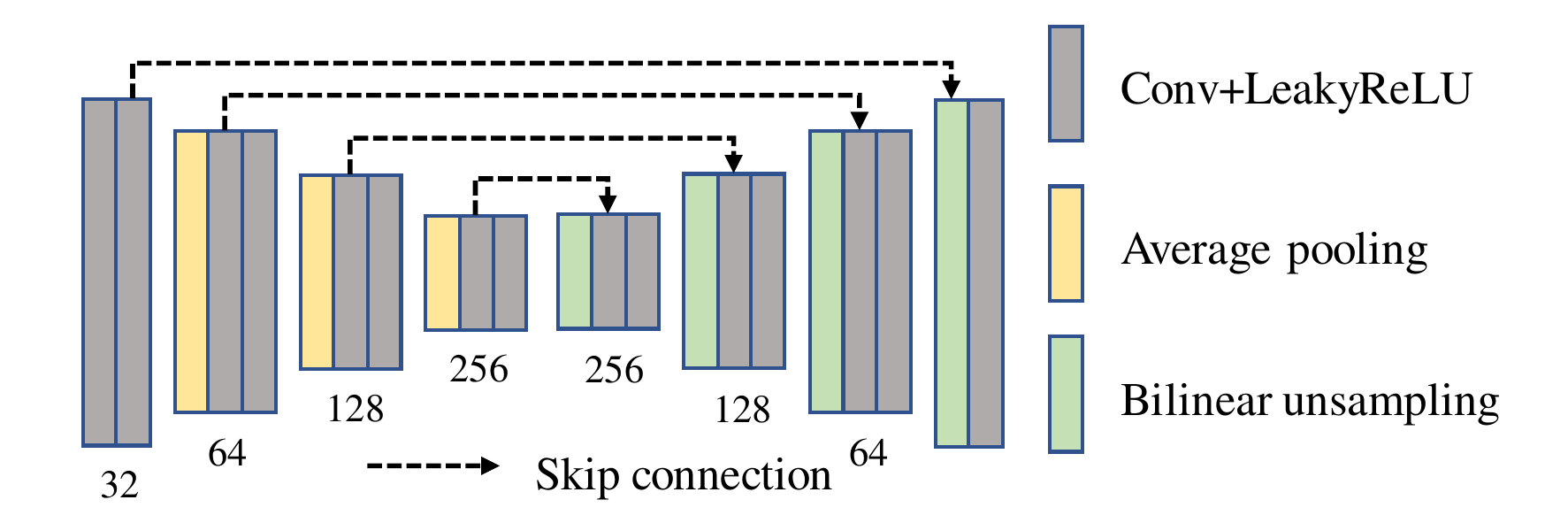}
\caption{The architecture our proposed Event-driven Optical Flow Mask Generation Network. The number below each block denotes the channel of corresponding output feature map.}
\label{fig:EDOMG_architecture}
\end{figure}

As shown in Fig.~\ref{fig:EDOMG_architecture}, our proposed Event-driven Optical Flow Mask Generation Network is a U-Net style network. The input is both event representation and the originals frames. The output is the mask of the bi-directional optical flow in the different direction in a anisotropic way. 

\subsection{The Refinement Network}
\begin{figure}
\centering
\includegraphics[width=\linewidth]{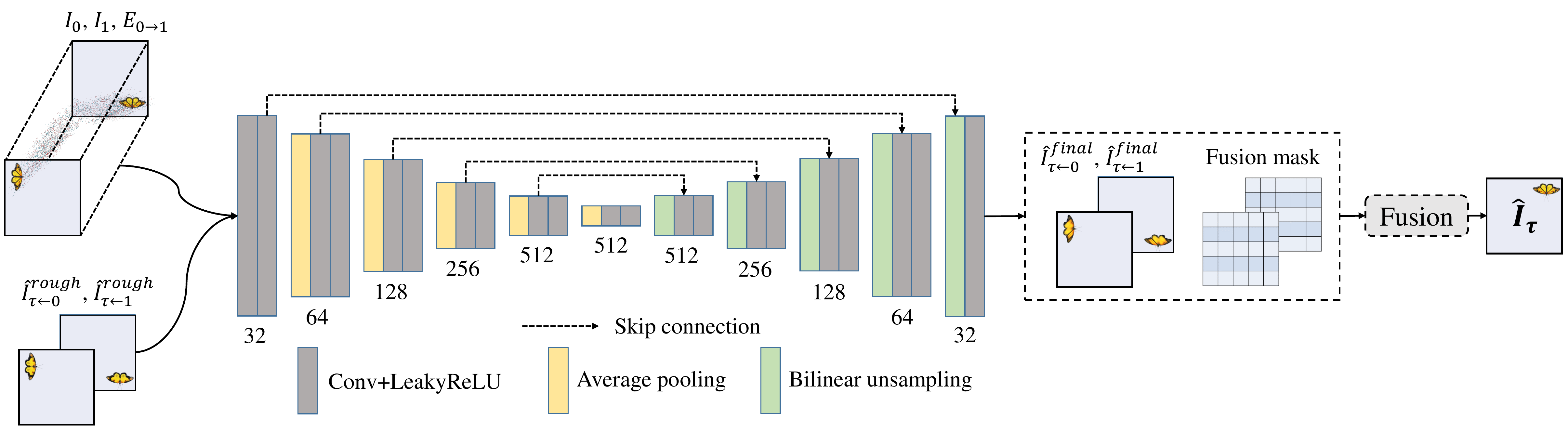}
\caption{The pipeline of refinement process. }
\label{fig:refinement}
\end{figure}

The architecture of our refinement network in proposed VFI model is shown in Fig.~\ref{fig:refinement}.  This refinement network is also a U-Net style network which is similar to that in \cite{SuperSloMo2018Jiang}.

\subsection{The Tailored IFNet}
\begin{figure}
\centering
\includegraphics[width=0.5\linewidth]{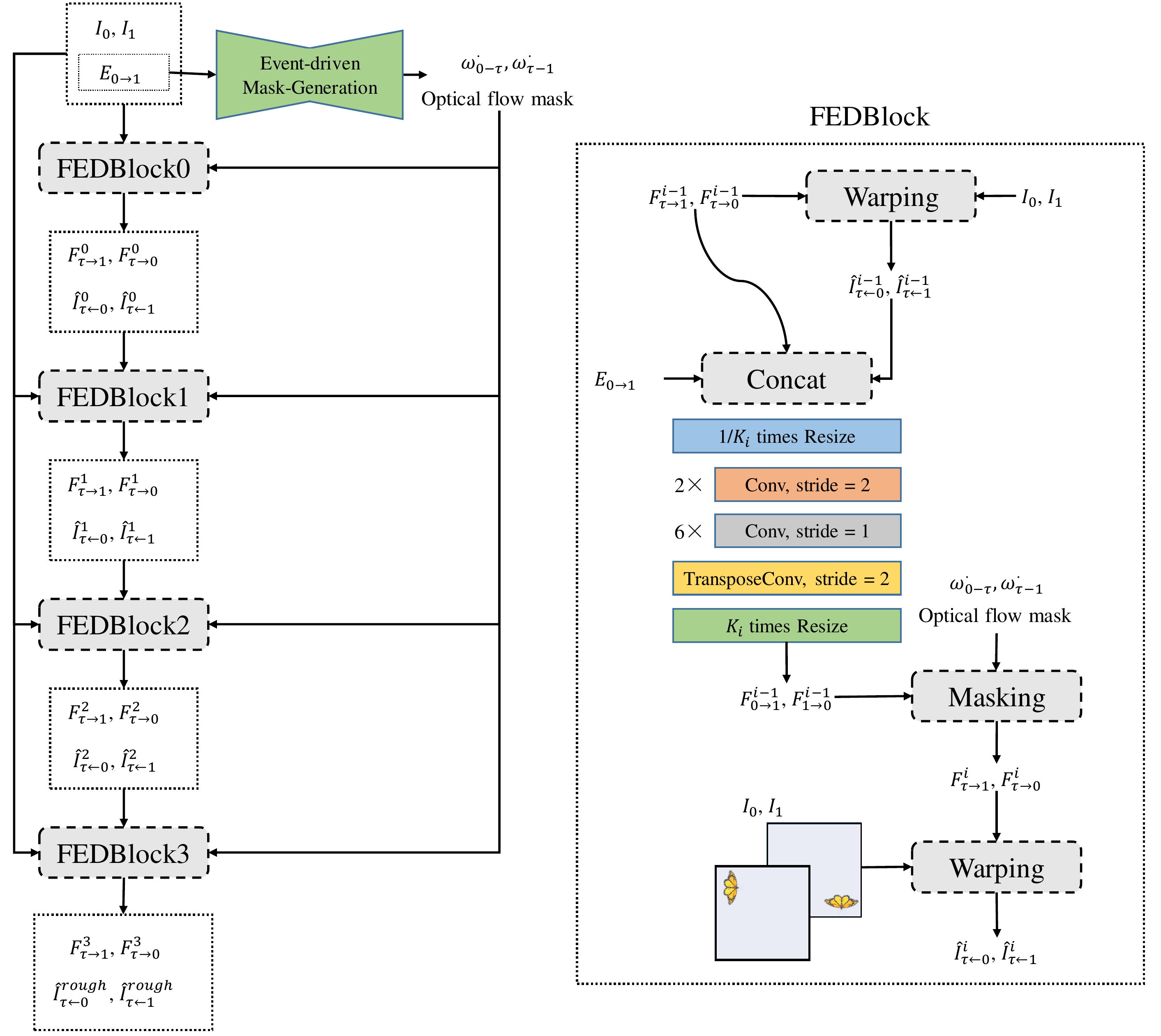}
\caption{The architecture of tailored IFNet.}
\label{fig:ifnet}
\end{figure}

As shown in Fig.~\ref{fig:ifnet}, we tailor the IFNet\cite{RIFE} to better model the motion with event-driven optical mask and event representation. 

\section{The Video Demo}
Here, we provide a demo to show the performance of our VFI model. The demos are provided in our submitted files. 

\clearpage

\end{document}